\documentclass[10pt]{article} 
\usepackage[preprint]{tmlr}

\usepackage{amsmath,amsfonts,bm}
\usepackage{graphicx} 

\def\eqref#1{equation~\ref{#1}}

\def\1{\bm{1}}

\DeclareMathAlphabet{\mathsfit}{\encodingdefault}{\sfdefault}{m}{sl}
\SetMathAlphabet{\mathsfit}{bold}{\encodingdefault}{\sfdefault}{bx}{n}

\usepackage{hyperref}
\usepackage{url}
\usepackage{graphicx}
\usepackage[export]{adjustbox}
\usepackage{booktabs}
\usepackage{makecell}
\usepackage{subfig}
\usepackage{bbding}
\usepackage{pifont}
\usepackage{wasysym}
\usepackage{amssymb}
\title{Leveraging LLMs for Predictive Insights in Food Policy and Behavioral Interventions}

\author{
    \name Micha Kaiser\thanks{Corresponding author} \email m.kaiser@jbs.cam.ac.uk \\
   \addr El-Erian Institute, Cambridge Judge Business School\\
    University of Cambridge 
   \AND
    \name Paul Lohmann \email p.lohmann@jbs.cam.ac.uk \\
   \addr El-Erian Institute, Cambridge Judge Business School\\
   University of Cambridge
    \AND
    \name Peter Ochieng \email po304@cam.ac.uk \\
     \addr YNOT Institute, Queen's College, Cambridge\\
    University of Cambridge
  \AND   
   \name Billy Shi\thanks{Currently at Apple} \email zs371@cam.ac.uk \\
   \addr Department of Engineering\\
    University of Cambridge 
   \AND
    \name Cass R. Sunstein \email csunstei@law.harvard.edu \\
    \addr Harvard Law School\\
    Harvard University 
   \AND       
   \name Lucia A. Reisch \email lr540@cam.ac.uk \\
    \addr YNOT Institute, Queen's College, Cambridge\\
     University of Cambridge
 }

\def\month{MM}  
\def\year{YYYY} 
\def\openreview{\url{https://openreview.net/forum?id=XXXX}} 

\begin{document}

\maketitle

\begin{abstract}
Food consumption and production contribute significantly to global greenhouse gas emissions, making them crucial entry points for mitigating climate change and maintaining a liveable planet. Over the past two decades, food policy initiatives have explored interventions to reshape production and consumption patterns, focusing on reducing food waste and curbing ruminant meat consumption. While the evidence of "what works" improves, evaluating which policies are appropriate and effective in specific contexts remains difficult due to external validity challenges. This paper demonstrates that a fine-tuned large language model (LLM) can accurately predict the direction of outcomes in approximately 80\% of empirical studies measuring dietary-based impacts (e.g. food choices, sales, waste) resulting from behavioral interventions and policies. Approximately 75 prompts were required to achieve optimal results, with performance showing signs of catastrophic loss beyond this point. Our findings indicate that greater input detail enhances predictive accuracy, although the model still faces challenges with unseen studies, underscoring the importance of a representative training sample. As LLMs continue to improve and diversify, they hold promise for advancing data-driven, evidence-based policymaking.


\end{abstract}

\section{Main}
 Food consumption and production, accounting for one-quarter to one-third of global greenhouse gas emissions, are key entry points for mitigating climate change, reducing mortality and morbidity, and keeping the world within liveable planetary boundaries  \citep{crippa2021food}. Over the past decade, food policy has been at the forefront of sustainability efforts, exploring various interventions to reshape production and consumption patterns and stimulate dietary shifts. These initiatives, including those targeting greenhouse gas emissions in the food system, have significantly advanced our understanding of which interventions are most effective in shifting individual behavior \citep{lohmann2024}. Today, climate scientists concur that the most effective approach is to focus on two critical behavioral strategies: reducing food waste and loss (FWL) \citep{gatto2024global,zhu2023cradle} and curbing ruminant meat consumption \citep{li2024reducing}. Just focusing on these two would lead to substantial strides in mitigating food-related emissions  \citep{geyik2023climate, crippa2021food, xu2021global, yang2024ghg}.
    
The challenge for policymakers aiming to mitigate climate change is to achieve significant and lasting behavior change. Various policy instruments are available to help drive such changes, from mandates and bans to financial incentives and disincentives, from information and education to behaviorally informed interventions using choice architecture \citep{ ammann2023review}. Evidence reviews on the effectiveness of such interventions guide policymakers towards the most effective yet acceptable policies \citep{lohmann2024, reisch2021mitigating}. Behaviorally informed food policies seem well-suited to be included in policy bundles since they can be effective, easy to implement, and often more acceptable to the public than hard policies like taxes and bans \citep{reisch2021shaping}. Moreover, they can be applied to a multitude of food system actors at individual and system levels (ibid.). With all these options at hand, how can policymakers evaluate which policies are the most appropriate and effective to tackle a specific policy problem (such as reducing food waste) in a particular setting (such as a public canteen run by a municipality in the UK) for a specific target group (such as vulnerable young consumers)? There are notable practical challenges as the effectiveness of intervention will vary depending on intervention levels, countries, and settings.
    
While the scientific literature and evidence synthesis networks increasingly produce helpful, systematic reviews and meta-regressions of hundreds or even thousands of studies, traditional approaches are resource-intensive, time-consuming, and hence quickly outdated in today's era of rapid data analysis \citep{egger2003important}. Systematic maps and reviews of case studies can inspire by showing the scope of possible policies. Still, they can hardly be generalized due to the heterogeneity of target populations and the power of context. Indeed, socio-economic, social, and cultural contexts and societal conditions, such as high or low trust in government, heavily influence the success of interventions; sometimes, the outcomes are not even in the predicted direction. In a nutshell, predicting policy outcomes is difficult, and external validity remains a serious challenge \citep{findley2021external}. 

While the idea of conducting mega studies seems to address some of these challenges, particularly concerning external validity \citep{milkman2021megastudies, duckworth2022guide}, the primary practical issues—namely, cost and time to conduct experimental studies—remain unresolved. Given the complexity and numerous factors influencing the success of interventions, there is a growing need to identify \textit{more efficient methods} for generating reliable predictions and supporting informed decision-making for researchers and policymakers. Artificial intelligence, particularly large language models (LLMs), offers new possibilities due to their ability to process vast amounts of data and identify patterns with much lower resource and time costs. Originally designed for text, LLMs now excel in various tasks, including coding \citep{coello2024effectiveness}, symbolic mathematics \citep{ahn2024large}, and scientific reasoning \citep{ai4science2023impact}.

There is increasing evidence that LLMs can be utilized for predictive tasks beyond their original design, such as stock market predictions \citep{lopez2023can} and even more general financial tasks \citep{wu2023bloomberggpt, xie2023pixiu, deng2023llms}. Additionally, LLMs have been employed to simulate complex daily behaviors and interactions—such as cooking or commuting— and may offer a realistic approach to agent-based modelling architectures \citep{park2023generative}. Moreover, LLMs have shown promising results in simulating classical economic, psycholinguistic, and social-psychological experiments, such as the \textit{Ultimatum Game} or the \textit{Wisdom of Crowds} \cite{aher2023using}. Horton \citep{horton2023large} further demonstrates how LLMs, functioning as computational representations of human decision-making (or "homo silicus"), can be integrated into economic and social science simulations by endowing the simulated agents with specific sets of information and preferences. Given these findings, it is unsurprising that there is a growing debate about whether, and to what extent, LLMs could replace humans in specific research tasks—particularly in psychological and behavioral studies in the near future  \cite{dillion2023can}—or be used to provide a more cost-effective adjustment of behavioral policies \cite{meng2024ai}.
A necessary condition for LLMs to simulate experiments correctly — or even whole survey responses \citep{namikoshi2024using} — is that previous experimental findings used for training are of high quality and replicable, ensuring they exhibit some external validity. To address this, statistical models  \citep{yang2020estimating} are increasingly employed to predict the replicability of experimental results in social science labs—an essential requirement for scientific progress in general  \citep{camerer2018evaluating}, specifically when using LLMs to predict the outcomes of specific policy interventions. These models have shown that factors such as sample size and effect sizes in the original studies strongly predict successful replication \citep{altmejd2019predicting}.
Despite these advances, there remain fundamental areas for improvement in the applicability of LLMs within the social sciences, including behavioral economics and cognitive psychology. These include a need for  broader access to high-quality research tools and a deeper understanding of the underlying social forces that guide behavior  \citep{bail2024can}. Demszky et al. \citep{demszky2023using} echo this sentiment, identifying critical gaps in the field, such as the lack of "keystone" datasets, standardized performance benchmarks, and shared computational infrastructure.

Here, we introduce PREDICT (Predictive Algorithm for Assessing External Validity and Identifying Contextual Tailored Interventions), an LLM-based decision support tool designed to address the foregoing challenges. Leveraging the ability to adjust a pre-trained LLM with context-specific and task-specific data (called fine-tuning \citep{howardruder2018finetune}), PREDICT offers new possibilities for predicting the outcomes of food-related policy interventions by processing vast amounts of data and identifying underlying patterns. As highlighted earlier, LLMs enable researchers to explore behavioral scenarios through simulations and uncover new insights, which can later be validated through real-world studies. Hence, we are exploring whether LLMs can accurately predict behavioral policy outcomes. We fine-tune a GPT-3.5 Turbo model on a comprehensive dataset of food-based interventions, encompassing 74 published scientific articles and over 200 effect sizes. Each experiment reports an average of 11,000 observations, resulting in a total of approximately 2.2 million observations measured across all included experiments. Additionally, we validate the predictive capabilities of our model by comparing predictions to 12 ongoing experiments that have not yet been published. We fine-tune the model using variations in prompt styles, dataset sizes, and prompt features, and find that the fine-tuned models can accurately predict statistical parameters, such as effect direction, correlation coefficients, and effect sizes, with effect direction accuracy reaching nearly 80\% in specific cases. Our findings shed light on how prompt style and training parameters affect prediction confidence, as well as the importance of specific dataset features in enhancing model performance. While LLM models continue to advance, our study contributes a significant proof of methodology in food policy research. We also aim to set a foundation for similar techniques applied in other policy research areas. In principle, LLM models could be used to identify potential solutions to a wide range of challenges, including highway safety, tobacco smoking, alcohol abuse, and take-up of beneficial programs.

\section{Results}
\subsection{LLM's Ability to Predict $r$ and $d$ for Empirical Behavioral Studies}
We report the ability of a fine-tuned GPT-3.5 turbo model to predict outcomes of empirical studies that measure dietary-based outcomes (e.g. food choices, sales, waste, etc.) due to behavioral interventions and policies. We compute the fine-tuned LLM model's fidelity in predicting multiple dependent outcomes simultaneously, each influenced by the same set of independent features. The outcomes to be predicted include the effect direction (positive or negative), the correlation coefficient (Pearson correlation, $r$) and the standardized effect size (Cohen's $d$). We report the statistical performance of LLMs across a collection of user studies in the dataset, specifically held out during the training process. We compare these predicted parameters with the empirical values reported in the source research paper and hence report accuracy in LLMs' predictions in terms of average \footnote{Critically, \textit{average} carries two layers of meaning: we compute the average absolute error over the test dataset, and then we compute this average absolute error 10 times for every fine-tuned model and report the error distribution due to the probabilistic nature of LLMs} absolute error, $\epsilon_{\mu}$ and variance in absolute errors, $\sigma^2_{\epsilon}$, for both \textit{r} and \textit{d}. 

\par There is mixed evidence regarding LLMs' capabilities in zero-shot prompting—i.e., prompting without any explicit fine-tuning. While some studies show that models like GPT-3 have demonstrated an emergent ability to find zero-shot solutions to a broad range of analogy problems  \cite{webb2023emergent}, others \cite{chang2024survey, liu2023pre} indicate that LLMs often fail to generate meaningful responses. This suggests that LLMs are not \textit{consistently} effective zero-shot reasoners for predicting human behavioral policy outcomes, such as those based on food interventions.

\begin{figure}[ht]
    \centering
    \includegraphics[width=1\linewidth]{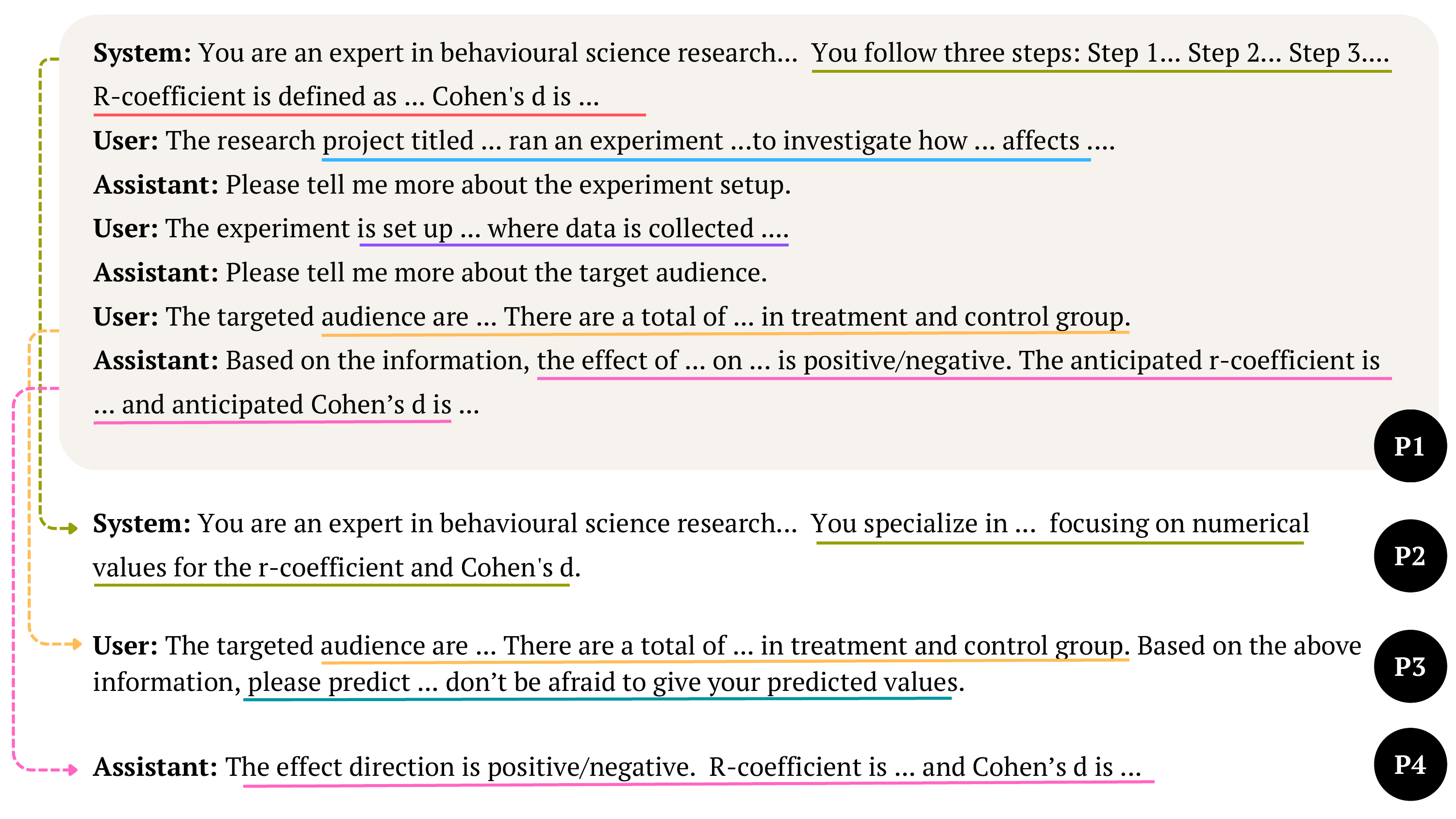}
    \caption{Prompt versions $P1$ to $P4$ showing conversational prompt templates used to fine-tune GPT-3.5-turbo models. $P1$ includes a system context message, research article summary, experiment setup, target audience and numerical results. Based on the $P1$ template, $P2$ removes step-by-step instructions; $P3$ adds guided completion; and $P4$ simplifies the numerical predictions.}
    \label{fig:prompts}
\end{figure}

LLM models fine-tuned on certain prompt formulations struggled to predict all or specific parts of the experimental results. In particular, we observed that these models found it more challenging to predict quantitative outcomes (e.g. $r$ and $d$ values) than qualitative ones. Therefore, for the four fine-tuned models ($MP_1$, $MP_2$, $MP_3$, $MP_4$), we report the probability of the model successfully making quantitative predictions, specifically the $r$ and $d$ values. The performance of these models is detailed in Table \ref{tab:prompt_models}, and the prompts used to fine-tune them are laid out in Figure \ref{fig:prompts}.

The results show that models $MP_1$ and $MP_2$, trained with verbose prompts, were more conservative in their predictions. Model $MP_1$ could not predict any of the $r$ and $d$ values, but it predicted the direction of the nudge 94.5\% of the time, with an accuracy of 36.7\%. Model $MP_2$ predicted the direction 68.2\% of the time with an accuracy of 23.1\%, and it predicted $r$ and $d$ values with a probability of 19\%. In contrast, the models $MP_3$ and $MP_4$ with less verbose prompts predicted the direction, $r$, and $d$ values 100\% of the time. This suggests that concise prompts elicit accurate numerical predictions from fine-tuned models more effectively. In particular, removing excessive context (e.g., the step-by-step instructions in $P2$) can help the LLM focus more on making numerical predictions as requested by the end of the prompt. We hypothesize that overly detailed prompts may lead the LLM to overthink and stray from aligning its responses with the training labels. Both $MP_3$ and $MP_4$ achieved 79\% accuracy in predicting the effect direction, and the improvement compared to $MP_2$ is likely due to adding guided completion ("Please predict ...") at the end of the prompt.

Regarding the prediction of $r$ and $d$ values, model $MP_3$ estimated $r$ with an average absolute error of -0.058 and $d$ with an error of -0.151, with variances of 0.142 and 0.441, respectively. Model $MP_4$ performed slightly better, predicting $r$ with an average absolute error of -0.009 and $d$ with an error of -0.051, with variances of 0.127 and 0.385, respectively. These results show that with effective prompt formulation, LLMs are few-shot learners \cite{brown2020language} for predicting human behavioral policy outcomes based on food policy interventions.

\begin{table}[ht]
\centering
\caption{Prediction performance of four fine-tuned LLM models, $MP_1$ to $MP_4$, fine-tuned with four different prompt formulations of the same dataset}
\label{tab:prompt_models}
\resizebox{\textwidth}{!}{%
\begin{tabular}{lccccc}
\hline
\multicolumn{1}{c}{Model} & \begin{tabular}[c]{@{}c@{}}Prediction Probability\\ for Effect Direction (\%)\end{tabular} & \begin{tabular}[c]{@{}c@{}}Effect Direction \\ Accuracy (\%)\end{tabular} & \begin{tabular}[c]{@{}c@{}}Prediction Probability\\ for $r$ and Cohen's $d$ (\%)\end{tabular} & \begin{tabular}[c]{@{}c@{}}Average Absolute \\ Error for $r$\end{tabular} & \begin{tabular}[c]{@{}c@{}}Average Absolute \\ Error for Cohen's $d$\end{tabular} \\ \hline
$MP_1$ & 94.5 & 36.7 & 0 & - & - \\ \hline
$MP_2$ & 68.2 & 23.1 & 19 & \begin{tabular}[c]{@{}c@{}}0.112\\ ($\sigma^2$ = 0.134)\end{tabular} & \begin{tabular}[c]{@{}c@{}}0.021\\ ($\sigma^2$ = 0.354)\end{tabular} \\ \hline
$MP_3$ & 100 & 79.0 & 100 & \begin{tabular}[c]{@{}c@{}}-0.058\\ ($\sigma^2$ = 0.142)\end{tabular} & \begin{tabular}[c]{@{}c@{}}-0.151\\ ($\sigma^2$ = 0.441)\end{tabular} \\ \hline
$MP_4$ & 100 & 79.0 & 100 & \begin{tabular}[c]{@{}c@{}}-0.009\\ ($\sigma^2$ = 0.127)\end{tabular} & \begin{tabular}[c]{@{}c@{}}-0.051\\ ($\sigma^2$ = 0.385)\end{tabular} \\ \hline
\end{tabular}%
}
\end{table}

\subsection{LLM's Prediction Accuracy Affected by the Number of Training Prompts}
To evaluate the sensitivity of an LLM's predictions to few-shot prompt fine-tuning \cite{mosbach2023few}, we fine-tune it on varying numbers of prompts ($N$). We begin by fine-tuning the model with 10 prompts (the minimum number OpenAI allows) and progressively increase the training data size to 144 prompts, the largest number of training prompts available, excluding the  held-out validation and test datasets. The performance of the fine-tuned models is shown in Figure 2. The best-performing models are those fine-tuned with 75 and 130 prompts, while models trained with 10 < $N$ < 75 and 75 < $N$ < 130 lose predictive power relative to the best-performing models.


\begin{figure}[h]
    \centering
    \includegraphics[width=1\linewidth]{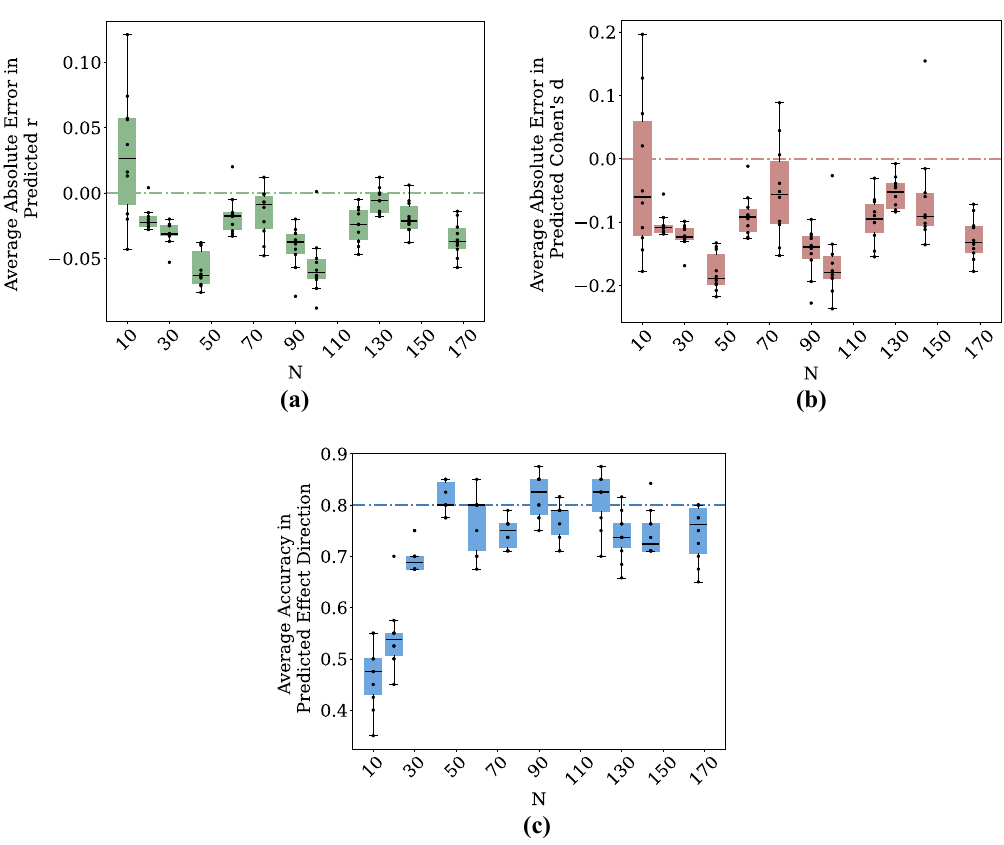}
    \caption{Plots of LLM's prediction performance versus varying number of training prompts, $N$. Namely, (a) average error in $r$ prediction vs. varying $N$, (b) average error in Cohen's-$d$ prediction accuracy vs. varying $N$, (c) effect direction prediction accuracy vs. varying $N$.}
    \label{fig:example}
\end{figure}

We hypothesize that the model's parameters are optimally fine-tuned with 75 prompts. As more data is added and training time increases, further adjustments to the LLM's parameters lead to a loss of previously acquired knowledge. The parameters are optimized again when the number of prompts reaches 130, but beyond this point, the model's predictive power begins to decline. To further demonstrate this hypothesis, we combine the 23 validation prompts (10\% of the dataset) with the training dataset, which brings the total number of training data to 167. This confirms our conjecture that adding training data examples beyond an optimal number may worsen the model's predictive performance. This pattern is well-known and likely related to the LLMs' tendency to inefficiently use longer contexts \cite{luo2023empirical}, which can lead to catastrophic forgetting and a subsequent loss of accuracy \cite{liu2024lost}.

\subsection{Prediction Accuracy Affected by Prompt Features}
We report the importance of various prompt features in the fine-tuning of an LLM by treating our prompt formulation as filling in a standard template with key parameters from empirical studies, such as the population type, geographical location, year of the experiment and sample size. This analysis aims to identify any uninformative features that might add noise and reduce the model's predictive accuracy, and hence we report the average predictive errors of feature-removed models bench-marked against the one with all features included (model $MP4$ as reported in Table 1). As shown in Figure 3, results show the only model that outperforms $MP4$ is the one fine-tuned without the title of the research article. This suggests that the titles introduce noise into the fine-tuning process as they sometimes carry information orthogonal to the causality between experimental interventions and measured outcomes; including this in the training prompt distracts the model and degrades its predictive performance. In contrast, all other features are demonstrated as essential for accurate target predictions. Notably, excluding the sample size leads to a significant drop in accuracy, highlighting its critical role in making precise predictions.

\begin{figure}[ht]
    \centering
    \includegraphics[width=1\linewidth]{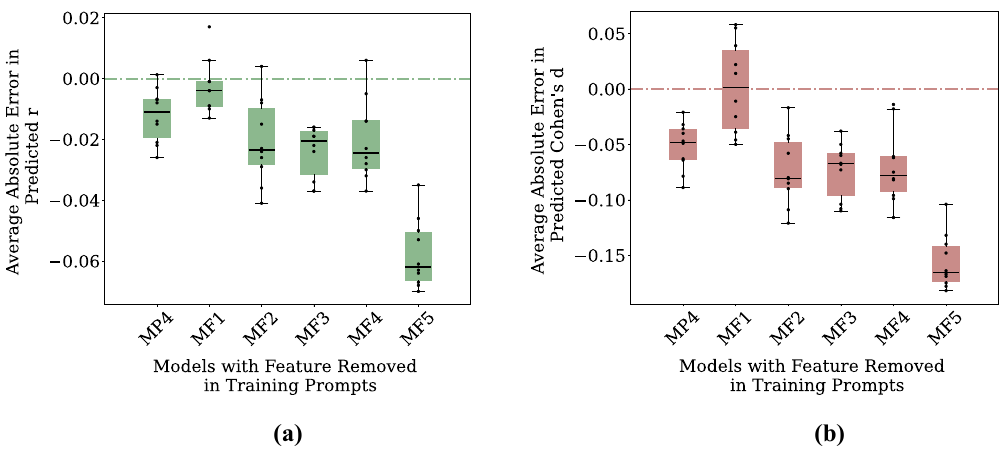}
    \caption{Plots of LLM's prediction performance for models with removed features. (a) average error in $r$ prediction (b) average error in Cohen's-$d$ prediction. Model \textbf{MP4} contains all features, as reported in Table 1. Model \textbf{MF1} has paper titles removed. Model \textbf{MF2} has geographic details removed. Model \textbf{MF3} has timeframe details removed. Model \textbf{MF4} has target audience details removed. Model \textbf{MF5} has sample size details removed.}
    \label{fig:enter-label}
\end{figure}

\subsection{Prediction Accuracy for Unseen Studies}
To further evaluate the predictive accuracy on experiments that were, with certainty, not used during the GPT-3.5 Turbo model training process, we extracted data from 12 experiments unpublished at the time of our model fine-tuning \cite{pizzo2024shapes, lohmann2024choice}. The main objective is to assess whether our fine-tuned models can generalize the knowledge acquired during fine-tuning to accurately predict future, unseen outcomes of specific policy interventions.
When applied to this test set, our best-performing model (fine-tuned with 130 prompts) predicts the effect direction, the $r$-coefficient, and Cohen's $d$ with 100\% coverage. On average, it correctly predicts the effect direction in approximately 6.6 out of 12 experiments, corresponding to an accuracy of 55\%. This is slightly below the 79\% accuracy observed in the baseline dataset but still higher than a naive estimator, which predicts the most common outcome ("negative") and achieves 50\% accuracy for this test sample.
The model shows an average error of -0.088 in predicting the $r$-coefficient and -1.368 in predicting Cohen's $d$. However, two experiments in the test data involve monetary incentives. Since only 5\% of the training and test datasets (i.e. 12 effect sizes) include monetary incentives, this small sample size may have limited the model's ability to learn the underlying relationship during fine-tuning, hence explaining the slightly weaker performance on these unseen cases. When we exclude these two experiments, the average error in the $r$-coefficient improves to 0.049, while the error in Cohen's $d$ decreases to 0.097.
Notably, these adjustments suggest enhanced predictive accuracy compared to a naive estimator, which would result in errors of 0.12 for the $r$-coefficient and 0.30 for Cohen's $d$.

\section{Discussion}
The results strongly suggest that LLMs hold significant promise as tools for policymakers, aiding in an exceptionally difficult task: selecting and evaluating behavioral policies and interventions. Specifically, the $MP_3$ and $MP_4$ fine-tuned models showed strong performance, achieving 79\% accuracy in predicting effect direction. Both models also produced highly accurate results with low variance in predicting correlation coefficients ($r$) and effect sizes (Cohen's $d$), indicating consistent performance. At the same time, we offer cautionary notes. A level of 21\% inaccuracy in predicting effect direction is a serious concern; it is important to find ways to meet that concern. In addition, the findings suggest that prompt formulation plays a critical role in fine-tuning LLMs. We find that including a high level of details in prompts can lead to diminishing returns, as the model may struggle to filter relevant information effectively. This insight can help refine best practices for prompt engineering \citep{ekin2023prompt}, particularly for tasks that require precise quantitative predictions. Our methodology also suggests the importance of iterative prompt tuning in achieving accurate LLM predictions.

Our analysis also indicates that more training data in the fine-tuning process does not necessarily lead to better model performance. In our case, the models fine-tuned with 75 and 130 prompts achieved the highest accuracy, outperforming those trained with either fewer or greater numbers of prompts. This suggests that even in presence of abundant data, there exists an optimal number (or a few optimal numbers) of training data examples to be included in model fine-tuning. When additional prompts are included, the model's performance initially decreases, likely due to over-adjustment and loss of previously acquired knowledge. The performance may improve again (in our case, with 130 prompts), but including further training data beyond this point again leads to a decline in accuracy, possibly due to over-fitting or parameter saturation. 

We also observed that the effectiveness of fine-tuning is highly sensitive to the features included in the prompt. Omitting crucial features significantly reduces the model's predictive power, while overloading the prompt with unnecessary information introduces noise, further diminishing accuracy. In this study, key features such as the year of the experiment, location, target participants, sample size, treatment group size, and control group size were essential for optimizing fine-tuning. However, including less relevant details, such as the research paper's title, added unnecessary noise and negatively impacted the fine-tuning process. Thus, careful consideration of feature selection in prompts is essential for achieving optimal model performance.

In addition to this, our analysis of the model's performance on previously unpublished experiments underscores the importance of a balanced and especially representative training dataset, particularly for enhancing predictive accuracy on unseen cases and further strengthening its utility for policy-relevant applications. As our findings indicate, carefully compiling a diverse range of interventions is essential for the model to capture all possible interaction effects that may emerge.

Our attempt to develop and test PREDICT as proof-of-concept offers significant promise for the future. These modest first steps seem to support the observation of an "unreasonable effectiveness of algorithms" in addressing public policy problems \cite{ludwig2024unreasonable}, a high potential upside. It is reasonable to speculate that our findings in the context of food policy might find parallels in other domains of behaviorally informed policy. As potential examples, consider efforts to improve road safety, to reduce smoking, to combat domestic violence, to promote energy conservation, to reduce unlawful immigration, to improve resilience against extreme heat, and to combat opioid addiction. LLMs might be enlisted to give policymakers and others greater clarity on what works and what does not. Despite initial effectiveness, the use of proposed models has some limitations. First, we have not tested these models in real-world scenarios at scale, and its effectiveness in practical applications remains uncertain. We expect significant variability in how different users might interact with a tool powered by these fine-tuned LLM models, and we have yet to identify which specific design features will be most supportive—or potentially problematic—in various contexts. Second, while the dataset used in this study includes 74 published papers, its coverage is not exhaustive: for example, we have not covered all geographical locations and target groups. Consequently, the dataset may not fully represent diverse populations, which could skew predictions and limit the generalizability of the model across different demographics and regions. The incomplete nature of the historical datasets could introduce biases in fine-tuning the LLM, potentially leading to suboptimal policy recommendations. As we present this promising methodology, we also take this opportunity to raise important questions on how LLM-assisted tools should (or should not) be used in policymaking settings, recognizing the convenience it brings by synthesizing historic experiments and the limitation and potential bias it might bring along.

Moving forward, the development of LLM-based research assistants like PREDICT for policymaking will need to address several critical questions:\\

\textbf{Building trust and confidence:} What criteria are necessary to establish trust and confidence in using such a tool? We hypothesize that a robust, scientifically independent database (e.g. empirical studies that have undergone rigorous validity and quality assessments) combined with an interdisciplinary team of authors and developers will be essential. The next step would involve real-world testing with policymakers to identify which features enhance or diminish trust in PREDICT. Co-design exercises and user interface copy-testing will be crucial to adapting the tool to meet the needs of policymakers in real-world political processes, thereby increasing its credibility and attractiveness.\\

\textbf{Assessing and mitigating risks:} What are the potential risks of using PREDICT as a decision-support tool, and how can these real and perceived risks be mitigated? Potential risks include the risk of error, technological challenges inherent in the use of LLMs, administrative and procedural barriers (such as the acceptability of AI-based tools in different policy contexts or regulatory frameworks), and ethical concerns. Addressing these risks will require tailored solutions, including transparency, education, and clear communication about the capabilities and limitations of such a research assistant. Trust in government and technology will also play a significant role in mitigating these risks.\\

\textbf{Interpreting LLM decisions:} Although fine-tuned LLMs can predict the effectiveness of policy interventions, they might be seen to do so in a `black box' manner, offering little insight into the rationale behind their predictions. This lack of interpretability may challenge policymakers who require transparent justifications for the recommendations. A key goal of this research is to develop techniques that make the decision-making process of LLMs more transparent and understandable, thereby increasing trust and usability.\\

\textbf{Evaluating long-term impact:}What are the opportunities for sustainable food policy beyond this initial proof-of-concept? How accurate are the predictions, and how much better can they be? Research has shown that government decision-makers often misjudge the benefits and costs of policy interventions. Ludwig et al. \citep{ludwig2024unreasonable}   argue that algorithms can improve the rank-ordering of policy options by marginal benefit, leading to better social outcomes. Future studies should compare PREDICT-based policies with traditional human-judgement-based approaches to determine the extent of improvement. This will require experimentation and long-term systematic monitoring of results, ideally in collaboration with policy units responsible for food policy.\\

\textbf{Exploring generalizability across domains:} While the fine-tuned LLMs demonstrate effectiveness in predicting outcomes related to behavioral interventions in food policy, it remains to be determined whether these models would perform similarly in other domains, such as transport, occupational safety, education, immigration, and environmental policy. Our findings with respect to a model trained solely on food policy data may or may not have parallels in these and other domains. Further research would be needed to assess the model’s ability to generalize across policy areas, ensuring that its predictive power remains robust in other contexts.\\

\section{Methods}
\subsection{Problem Formulation}
Given a collection of empirical studies extracted from published research papers that report dietary-based outcomes influenced by food behavior policies and interventions, our objective is to extract a set of experiment descriptions $\mathcal{X}$ and their associated outcomes $\mathcal{Y}$. We represent the features of an experiment as $x \in \mathcal{X}$ and associate them with their respective outcomes $y \in \mathcal{Y}$. The dataset, comprising all described experiments and their outcomes from the research papers, is defined as $\mathcal{D} = {(x_1, y_1), \dots, (x_N, y_N)}$. Our goal is to fine-tune a large language model using a subset $\mathcal{D}_{training}\subset \mathcal{D}$, thereby establishing a function $f: \mathcal{X} \rightarrow \mathcal{Y}$ such that $f(x)$ accurately approximates $y$ over a distribution of inputs $x$. Once the fine-tuned LLM $f$ has been established, Another data subset $\mathcal{D}_{test}\subset \mathcal{D}$ is  used to evaluate how well it estimates the distribution of $\mathcal{Y}$ given $\mathcal{X}$.

\par Concretely, the methodology seeks to evaluate the ability of LLMs to predict an outcome of food behavioral policies and interventions accurately. We assess this by first fine-tuning an LLMs using a training dataset containing a description of existing behavioral policies and interventions with already known outcomes. The fine-tuned LLM is then evaluated on test data. We design an algorithm that takes as its input an experiment's important details (features) and uses these details to construct a prompt that queries an LLM. The prompt consists of typical information that describes food human behavioral experiments such as the interventions, the year the experiment was conducted, the location of the experiment, targeted participants, sample size, treatment group size and controlled group size. Based on the query prompt, the LLM predicts a probability distribution over the space of all possible numerical values that evaluate an intervention's effect on food consumption patterns.\\
\subsection{Prompt generation} 
LLMs are autoregressive models that generate predictions sequentially by estimating the probability distribution of the next token in a sequence, i.e., $p(t_i | t_1, t_2, \cdots, t_{i-1})$ for any token $t_i$ given previous tokens ${t_1, \cdots, t_{i-1}}$ \cite{aher2023using}. To elicit responses from LLMs, prompts are carefully designed to obtain the best generative response. Prompts can generally be categorized into two main types: cloze prompts \cite{petroni2019language} \cite{cui2021template}, which fill in blanks within a textual input, and prefix prompts \cite{li2021prefix} \cite{lester2021power}, which complete a string prefix. The choice of prompt type depends on the specific task and the LLM model being used. In our case, we adopt the prefix prompt.

To construct the prompt $p$, we developed a prompt constructor module, $f_{\text{prompt}}(t, x)$, which takes a manually crafted prompt template $t$ (with some slots left blank) and a textual description $x$ of human behavioral experiments on food. This module automatically extracts key details from the description $x$ and fills in the missing slots in the template, thereby generating a complete prompt $p$, i.e., $p = f_{\text{prompt}}(t, x)$.
Designing an effective prompt template is essential for both fine-tuning and querying the LLM \cite{sclar2023quantifying}. The LLM's performance is particularly sensitive to the wording, feature selection, and format of the prompt \cite{sclar2023quantifying}. To address this, we experimented with various versions of the prompt template to identify the one that produces the best results on the test dataset.

In our case, the features included in the prompt were fixed based on experimental descriptions extracted from empirical studies on human food behavior. We ensured that all prompt versions contained the same features to avoid any spurious bias from omitting key information. Since we used the OpenAI GPT-3.5 turbo LLM\footnote{\url{https://platform.openai.com/docs/models/gpt-3-5}} via the chat completion API, the format of the prompt was also constrained. OpenAI's LLMs are trained to accept input formatted as a conversation, where the messages parameter contains an array of message objects organized by role (see an example in Figure 1). We adhered to this chat completion format.
\par Once a prompt $p$ has been established, it serves as a seed for generating additional candidate prompts. Starting with the seed prompt $P_1$, we generated three more candidate prompts, $P_2$, $P_3$, and $P_4$, through iterative paraphrasing. Each subsequent version was derived from the previous one; for example, $P_3$ was developed by adding a completion guide sentence at the end of $P_2$. We ensured that each new candidate prompt was less verbose compared to its parent prompt, which we found to improve to model performance.

\subsection{Dataset}
 The dataset was initially compiled using systematic search procedures  \cite{lohmann2024}. It included 74 published research papers that evaluated food policy interventions to shift individuals towards more sustainable food consumption or reduce food waste. The search was conducted across several scientific bibliographic databases: Web of Science Core Collections Citation Indexes, Scopus, MEDLINE and Google Scholar (first 20 pages of the output).
 
 The final sample was limited to quasi-experimental and experimental studies that measured the actual (or incentivized) behavior of individuals or households and provided valid counterfactuals to quantify intervention effects. Any type of intervention (e.g. monetary, information provision, or behavioral nudges) was considered. Details on the exact inclusion and exclusion criteria are provided in \cite{lohmann2024}. 
 
From each research paper, we manually extracted key details of the experiments, including the title of the experiment, a brief description of its goal, the location or region where the study was conducted, the targeted population, the total sample size, the sizes of both the treatment and control groups, and the results of the experiment. Specifically, we captured the direction (positive or negative) and magnitude of the effect induced by the intervention. These effect sizes were subsequently converted into standardized effect size measures, including the $r$-coefficient and Cohen's $d$ using the standard formulae described in \cite{ringquist2013meta}. In some instances, research papers contained multiple experiments, outcomes or interventions. In total, the 74 papers yielded 208 individual effect sizes, which were used to generate 208 distinct prompts. These prompts were then split into 144 for training, 23 for validation, and 41 for testing. 

Additionally, we calculated the $r$-coefficient and Cohen's $d$ for 12 effect sizes from ongoing food-related experiments, whose results had not been published at the time of the training and fine-tuning of the GPT-3 model. All experiments were preregistered, and we included these results as an additional validation check to rule out the possibility that the model had prior knowledge of the predicted outcomes.

\subsection{Fine-tuning}

Traditionally, leveraging pre-trained models involved gradient-based fine-tuning on downstream tasks, using pre-trained parameters as an initialization step \cite{ding2023parameter}. However, as LLMs have scaled to hundreds of billions of parameters, they have demonstrated properties that facilitate few-shot learning, making them adaptable to specific tasks with minimal examples \cite{brown2020language}. This process, known as few-shot fine-tuning, involves training the LLM on a small, task-specific dataset to adjust its parameters for enhanced performance on a particular task \cite{vm2024fine}.

For this study, we fine-tuned the LLM using a dataset of 144 training prompts and 23 validation prompts. Each prompt included a comprehensive description of an experiment's features, such as the experiment's title, the nudge applied to the treatment group, the location or region where the experiment was conducted, the targeted population, the total sample size, and the sizes of both the treatment and control groups. Additionally, each prompt included the results of the experiment, detailing the direction of the effect (positive or negative) induced by the nudge, along with labelled statistical metrics like the $r$-coefficient and Cohen's $d$.

These prompts were used during the fine-tuning process to adjust the LLM's internal parameters, thereby improving its ability to predict the impact of various nudges on food behavior. By incorporating both training and validation prompts, we aimed to ensure that the model could generalize well to unseen data while maintaining high performance on the specific task of understanding and predicting behavioral outcomes based on experimental details. The hyper-parameters (e.g. learning rate) are chosen automatically by the OpenAI completions API.

\par Each version of the prompt formulations ($P_1$, $P_2$, $P_3$, and $P_4$) was used to generate a corresponding fine-tuned model ($MP_1$, $MP_2$, $MP_3$, and $MP_4$).
\subsection{Inference}
To evaluate the accuracy of the predictions generated by the fine-tuned LLM models, we assess their ability to predict the correct effect direction, Pearson correlation coefficient ($r$), and Cohen's $d$ values induced by the nudge. The prediction accuracy of the LLM model is determined by first calculating the absolute error between the LLM's predicted values and the corresponding reported values from the literature. For example, if the model predicts $r = -a$ and the actual reported value is $r = q$, the absolute error is calculated as $| -a | - | q |$. A positive absolute error indicates that the model overestimates the effect, while a negative error suggests underestimation. Then, we compute the mean across all 41 test data examples, before re-running the LLM inference 10 times and report a distribution of the average absolute errors.

\bibliography{bibi}
\bibliographystyle{tmlr}
\end{document}